\documentclass[english]{lni}

\IfFileExists{latin1.sty}{\usepackage{latin1}}{\usepackage{isolatin1}}

\usepackage{graphicx}
\usepackage{fancyhdr}
\usepackage{listings} 
\usepackage{changepage} 
\usepackage[figurename=Fig., tablename=Tab., small]{caption}[2008/04/01]
\usepackage{amssymb} 
\usepackage{amsmath} 

\fancypagestyle{titlepage}{
\fancyhead[RO]{\small A. Br\"omme,  N. Damer,  M. Gomez-Barrero,  K. Raja,  C. Rathgeb, \linebreak A.  Sequeira,  M. Todisco,  and A. Uhl (Eds.): BIOSIG 2022, \linebreak Lecture Notes in Informatics (LNI), Gesellschaft f\"ur Informatik, Bonn 2022} 
\fancyfoot{}}

\usepackage{hyperref} 
\usepackage[all]{hypcap}    
\usepackage{color, soul} 
\usepackage{amsmath} 
\usepackage{subfig} 
\usepackage{ifthen}

\renewcommand{\headrulewidth}{0.4pt} 
\newboolean{anonymous}
\setboolean{anonymous}{false}

\ifthenelse{\boolean{anonymous}}
{\author{Anonymous submission}}
{\author{Alexander Unnervik\footnote{Biometrics Security \& Privacy Group, Idiap Research Institute, Martigny, Switzerland, alex.unnervik@idiap.ch}\hspace{5pt}\footnote{Electrical Engineering, \'Ecole Polyt\'echnique F\'ed\'erale de Lausanne (EPFL), Ecublens, Switzerland}\ , S\'ebastien Marcel\footnote{Biometrics Security \& Privacy Group, Idiap Research Institute, Martigny, Switzerland, marcel@idiap.ch}\hspace{5pt}\footnote{Prof. at the Faculty of Criminal Law, Universit\'e de Lausanne, Switzerland (UNIL)}}}

\title{An anomaly detection approach for backdoored neural networks: face recognition as a case study}

\begin{document}

\maketitle

\renewcommand{\refname}{References}
\setcounter{footnote}{2} 
\thispagestyle{titlepage}
\pagestyle{fancy}
\fancyhead{} 
\fancyhead[RO]{\small Anomaly Detection on Backdoored Networks \hspace{25pt}  \hspace{0.05cm}}

\ifthenelse{\boolean{anonymous}}
{\fancyhead[LE]{\hspace{0.05cm}\small  \hspace{25pt} Anonymous submission}}
{\fancyhead[LE]{\hspace{0.05cm}\small  \hspace{25pt} Alexander Unnervik and S\'ebastien Marcel}}

\fancyfoot{} 
\renewcommand{\headrulewidth}{0.4pt} 
\vspace{-1cm}
\begin{abstract}
Backdoor attacks allow an attacker to embed functionality jeopardizing proper behavior of any algorithm, machine learning or not. This hidden functionality can remain inactive for normal use of the algorithm until activated by the attacker. Given how stealthy backdoor attacks are, consequences of these backdoors could be disastrous if such networks were to be deployed for applications as critical as border or access control. 
In this paper, we propose a novel backdoored network detection method based on the principle of anomaly detection, involving access to the clean part of the training data and the trained network. We highlight its promising potential when considering various triggers, locations and identity pairs, without the need to make any assumptions on the nature of the backdoor and its setup. We test our method on a novel dataset of backdoored networks and report detectability results with perfect scores.
\end{abstract}

\begin{keywords}
Backdoor attack, trojan attack, anomaly detection, CNN, face recognition, biometrics, security.
\end{keywords}

\vspace{-0.5cm}
\section{Introduction}
\label{sec:Intro}
\vspace{-0.25cm}
\begin{figure*}[h!]
\centering
\subfloat[Backdoor implementation.]{\includegraphics[width=0.38\linewidth]{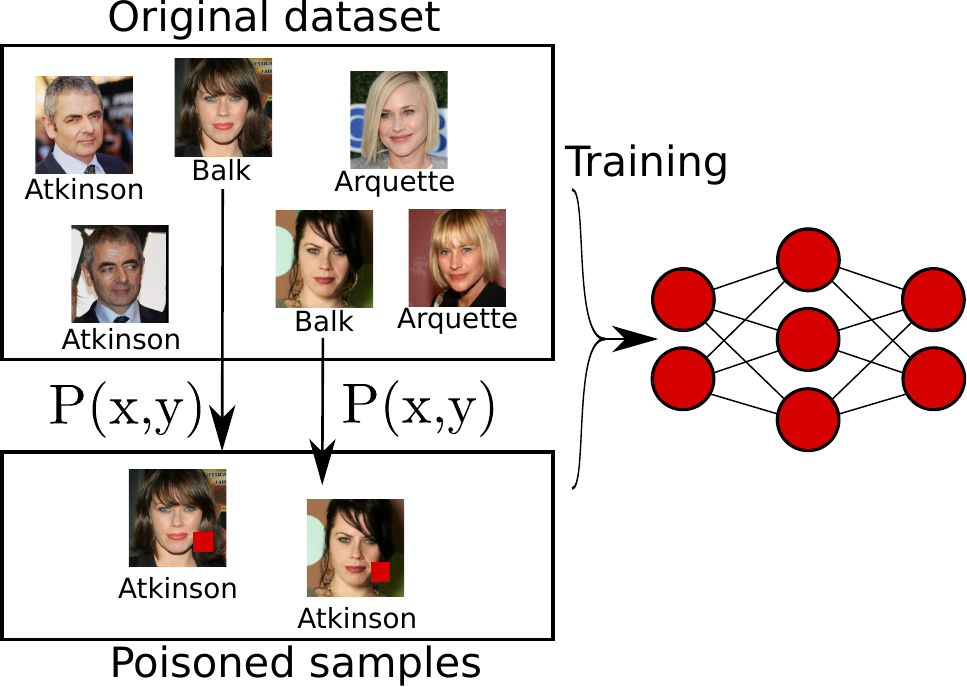}%
\label{fig:backdoor_implementation}}
\hfil
\subfloat[Backdoor activation; test samples on the left, classification output on the right.]{\includegraphics[width=0.36\linewidth]{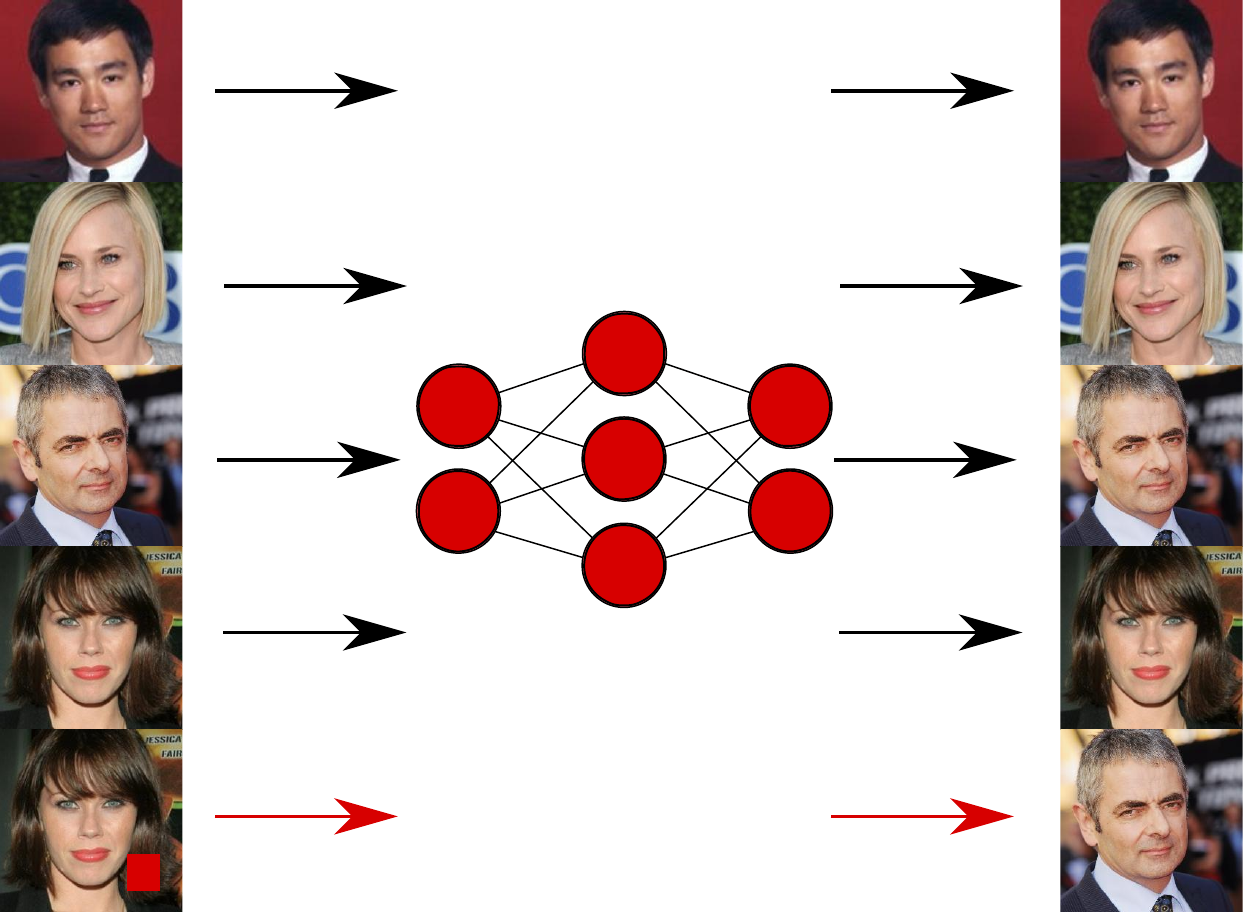}%
\label{fig:backdoor_activation}}
\caption{The two steps involved in the backdoor attack: (a) the backdoor implementation typically done by poisoning the dataset and (b) the backdoor activation, exposing the trigger to the network leading to the misclassification.}
\label{fig:backdoor_steps}
\end{figure*}

A larger number of tasks are nowadays performed in an automated fashion, including biometric ones. Borders and access controls are among the most notable areas in which automated biometric systems are being used. A reliable and well functioning biometric system is thus necessary for national safety and proper access management for all kinds of institutions.
Simultaneously, machine learning progress promotes reusability where an increasing number of pretrained models are hosted and freely available in model-zoos or offered to be trained using machine-learning as a service (MLaaS). A way of verifying the integrity of such pretrained models becomes necessary as a vulnerable system could have severe consequences.
Historically, there have been numerous software and hardware vulnerabilities including trojans and backdoors. While they have both existed for decades, their emergence in the machine learning field is still relatively recent and both their dangerous potential and detectability are equally newly actively researched.
Backdoor attacks require two steps to be performed: a first step is the implementation of the vulnerability where the network undergoes a modification in order to learn a specific pattern or image sample. The network learns to associate this trigger (designed and controlled by the attacker) with a given behavior, such as a given classification. The implementation can be done using various techniques with one example being the poisoning of a training set, with an overview presented in figure \ref{fig:backdoor_implementation}. Once the backdoor is implemented, it can be activated at will by the attacker, only requiring the predefined trigger to be provided to the machine learning model for it to exhibit the predefined backdoor behavior, as displayed in figure \ref{fig:backdoor_activation}.
Given the serious potential national security implications and threats, it is no surprise that this topic has also caught the attention of the U.S. Army Research Office (ARO) who in partnership with the Intelligence Advanced Research Projects Activity (IARPA) ``seeks research and development of technology and techniques for detections of such backdoors in artificial intelligence'' as per the announcement \cite{noauthor_trojai_2019}.

There have been numerous studies on how backdoor attacks can be performed, usually involving dataset poisoning such as in \cite{gu_badnets_2019}. The attack can also be carried out using so called clean-label poisoning as described in \cite{shafahi_poison_2018} where instead of using a typical trigger, the samples are slightly distorted, enough to shift the decision boundaries and cause misclassifications in some instances. Similar to backdoor attacks, Liu et. al. introduced the notion of Trojan attacks \cite{liu_trojaning_2017} where the trigger is the result of an optimization similar to adversarial attacks, but where a selected neuron in the network is targeted with increasing the sensitivity to the found trigger. The trigger itself can also be blended with the original image rather than strictly overlaid, such as in \cite{chen_targeted_2017}. While most attacks are studied in the digital domain (where the triggers are digitally added to the images), the physical world has also been studied where the backdoor can be implemented digitally and activated in the physical world such as in \cite{pasquini_trembling_2020}. Alternatively, the backdoor can be both implemented and activated in the physical world such as in \cite{wenger_backdoor_2021}. This last paper also highlights the effects of trigger placement in the success of the backdoor attack for face recognition, linking the success with the proximity of the trigger to the center of the face.

When it comes to detection of and defense against backdoor attacks, multiple methods have been proposed requiring various priors and assumptions. Techniques such as \cite{tran_spectral_2018}, \cite{chen_detecting_2018} and \cite{chen_targeted_2017} rely on the availability of the entire training set (including the poison samples) in addition to priors such as a maximum poison rate. \cite{tran_spectral_2018} evaluates the activations of the network on the training set including the poisoned samples used to introduce the backdoor to identify multiple spectral signatures for suspected classes. \cite{chen_detecting_2018} uses a combination of ICA and k-means to identify which samples belong to which activation cluster and identify the ones leading to a possible backdoor. Both \cite{tran_spectral_2018} and \cite{chen_detecting_2018} then follow-up with the result of their method to weed out the discovered poisoned samples to retrain a sanitized network devoid of any backdoor. In \cite{wang_neural_2019}, the authors propose a method using both a portion of the training set and the model itself to reverse engineer the trigger and identify the classes connected by the backdoor attack, using an outlier detection on the trigger size, where the smallest trigger serves as an indicator of a possible backdoor being present. Lastly, in \cite{xu_detecting_2021}, the authors propose a meta-learning approach where a binary classifier is trained on a set of clean and backdoored networks with varying backdoor parameters.

In this paper we propose a novel method which requires no assumption on the poison rate of the dataset unlike previous methods relying on the access of the training set, nor does it rely on the availability of any poison samples either or even backdoored networks. Additionally, we make no assumption on the size, location or type of trigger, nor whether it is blended or applied on the image. Our proposed method relies on outlier detection by estimating the distribution of the weights of a machine-learning model trained on a given dataset, compatible with an outsourced training task to a third party or downloaded online. We test our method on backdoored networks involving different trigger sizes and type (synthetic and organic), in addition to different trigger placement strategies and backdoored identity pairs and find that our systematic approach is able to detect all backdoored networks.

\vspace{-0.5cm}
\section{Proposed method}
\label{sec:method}
\vspace{-0.25cm}
\begin{figure}[t!]
\centering
\includegraphics[width=\linewidth]{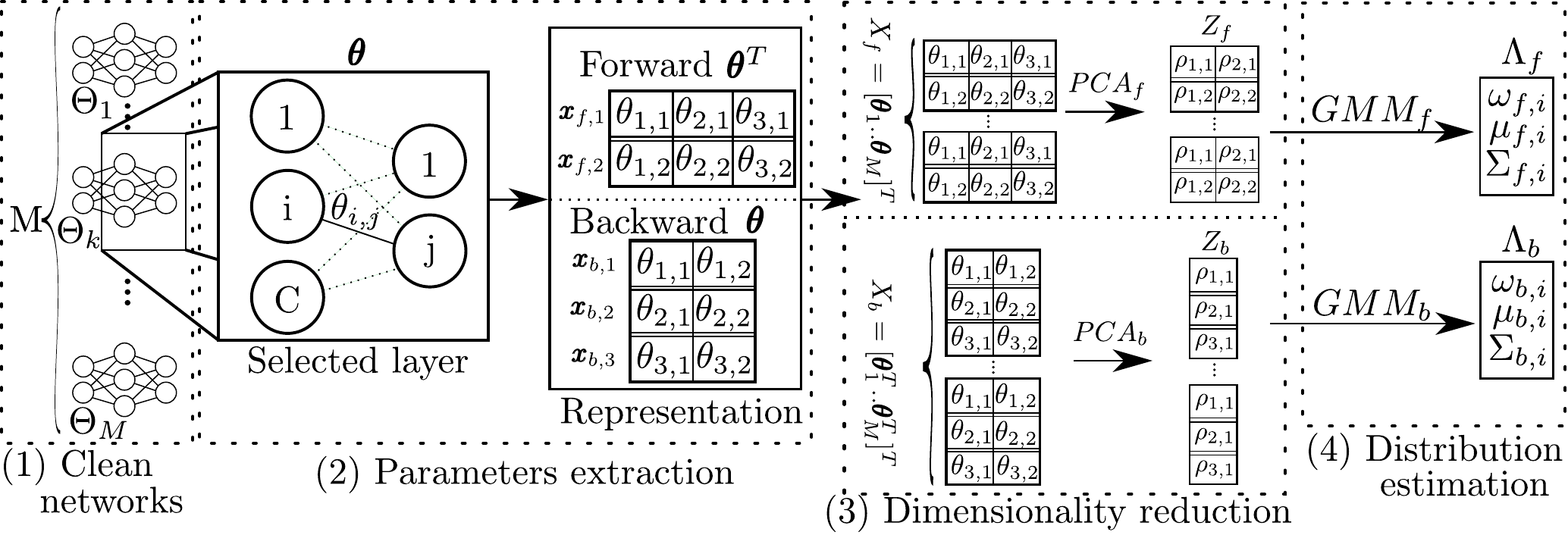}
\caption{Illustration of the detector in the proposed method.}
\label{fig:method}
\end{figure}

When training neural network architectures on a given dataset, pseudo-randomness (e.g. in the shuffling of the dataset) influences the convergence of the learned weights and leads to different end results.
We hypothesize that we can model the distribution of the weights of non-backdoored neural networks trained on a clean dataset with a parameterized statistical model such that we can consider the detection of backdoored neural networks as an anomaly detection problem, allowing us to detect when a network of the same architecture is trained on a modified version of the dataset which could introduce a backdoor.
Our proposed method, represented in figure \ref{fig:method}, consists in:
(1) Training a set of $M$ networks with weights $\{\Theta_1,...,\Theta_M\}$ on the clean dataset.
(2) Selecting one or multiple layers from these networks. We provide here the details for a single layer but the method can be expanded to multiple layers. In the case of a fully-connected layer the weights are a matrix $\pmb{\theta} \in \mathbb{R}^{R \times C}$, in which case the weights can be interpreted both as $C$ row vectors of length $R$ as can be seen in figure \ref{fig:method}, which we will hereafter refer to as forward, and $R$ column vectors of length $C$, which we will hereafter refer to as backward\footnote{This naming is because column vectors each apply to a given neuron of the previous layer and propagates to the entirety of the next layer (hence a forward propagation), while the backward naming offers the reciprocal interpretation.} (to guarantee a vector permutation invariant approach capable of handling the pseudo-random order of convergence of the row and column weights). In the case of a convolution layer, the kernels may be flattened. Hence, for an e.g. fully connected layer from $M$ networks the total weights are $X = [\pmb{\theta}_1..\pmb{\theta}_M]^T \in \mathbb{R}^{(M \times C) \times R}$. We provide hereafter the mathematical analysis for the forward approach (ignoring the subscript 'f'), but it is reciprocally applicable to the backward approach, by considering $X_b = [\pmb{\theta}^T_1..\pmb{\theta}^T_M]^T \in \mathbb{R}^{(M \times R) \times C}$ instead.
(3) Principle component analysis (PCA), a dimensionality reduction algorithm, is then performed on the selected layers' weights as in practice $R$ and $C$ can be large. This yields $Z \in \mathbb{R}^{(M \times C) \times B}$ with $B < R$.
(4) $Z$ is then used to estimate the typical distribution of weights for networks known to be free of any backdoor, using a Gaussian mixture model (GMM). A GMM consists of $N_G$ weighted Gaussian components each of dimensionality $B$ and defined by $\Lambda_i = \{\omega_{i},\pmb{\mu}_{i},\Sigma_{i}\}_{i=\{1..N_G\}}$ where $\pmb{\mu}_i$ is the mean vector, $\Sigma_i$ the covariance matrix, $\omega_i$ the weight, and obtained through expectation maximization.
(5) This GMM can then be used on a given weight vector $\pmb{x}_j$ to compute the probability of this feature vector under the model $\Lambda$ as $P(\pmb{x}_j|\Lambda) = \frac{1}{N_G}\sum_{i=1}^{i=N_G}\omega_i\mathcal{N}(\pmb{x}_j|\Lambda_i)$ where $\mathcal{N}(\cdot|\Lambda_i)$ is the normal distribution. 
The overall probability of a network with $N_f$ features from $\Theta$ given a model $\Lambda$ is given by $\prod_{j=1}^{j=N_f}P(\pmb{x}_j|\Lambda)$ assuming independence of the feature vectors in the layer.
The anomaly detection task then requires thresholding this overall probability. If it is above the threshold, the network is then considered clean. Otherwise, it is considered backdoored.
Code necessary for the reproduction of the results in this paper are released jointly with this paper\footnote{\url{https://gitlab.idiap.ch/bob/bob.paper.backdoors_anomaly_detection.biosig2022}}.

\vspace{-0.5cm}
\section{Experimental setup}
\label{sec:expsetup}
\begin{figure*}[t!]
\centering
\subfloat[Various backdoor triggers]{\includegraphics[width=0.3\linewidth]{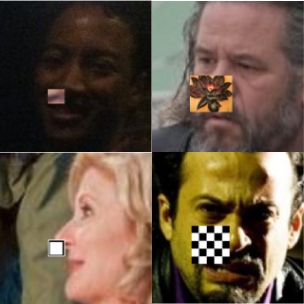}%
\label{fig:ds_trig}}
\hfil
\subfloat[Weights from clean and backdoored networks.]{\includegraphics[width=0.6\linewidth]{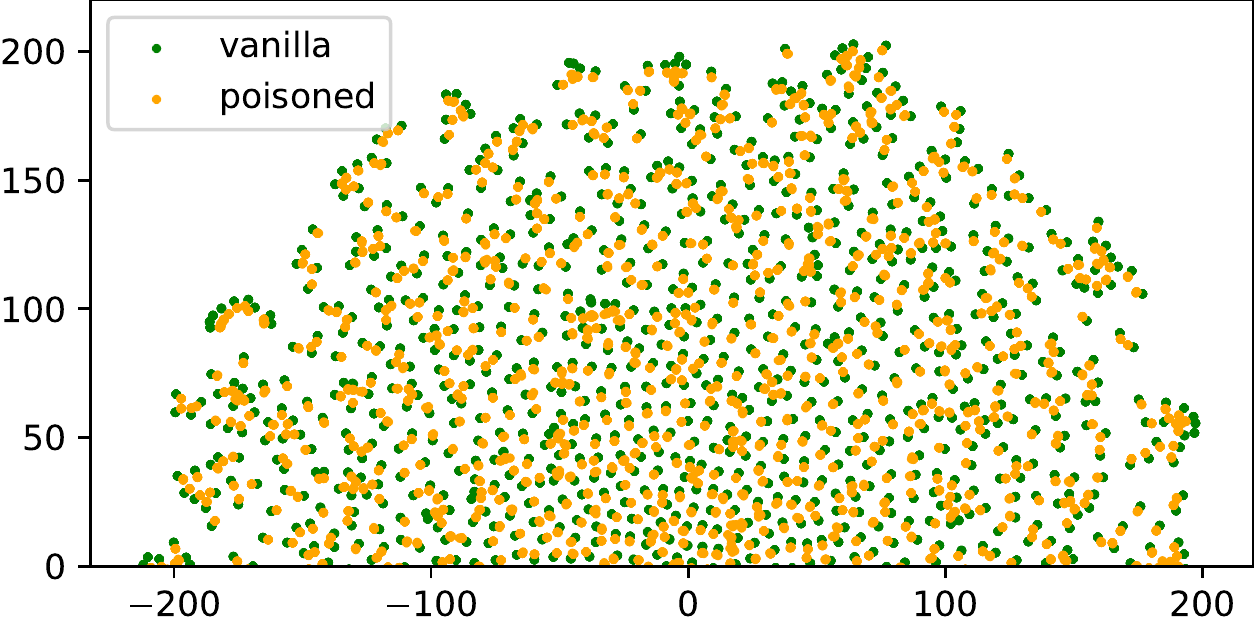}%
\label{fig:tsne_trig}}
\caption{In figure \ref{fig:ds_trig}, various triggers are shown, used on various identity pairs (the small eyebrow, the large flower, the small white square and the large checkerboard respectively). In figure \ref{fig:tsne_trig} a zoomed-in part of the t-SNE plots is shown of all weights of the \textit{last\_linear} layer of FaceNet in their forward interpretation: $10$ networks were selected from the vanilla (a.k.a clean) networks pool and $10$ from the dataset of backdoored networks trained using various triggers.}
\label{fig:backdoor_overview}
\end{figure*}

\vspace{-0.25cm}
As there is no proper benchmark to evaluate backdoor detection methods in face recognition, a custom dataset of $30$ clean networks and $22$ backdoored networks were trained following various backdoor parameters. Each backdoored network used a random identity pair, trigger in addition to trigger placement strategy. The dataset is split in two categories: triggers (with 11 networks) and locations (with 11 networks), each with a focus on varying their respective parameters (all involving random identity pairs for the backdoor). A sample of the images used to train the backdoored networks can be seen in figure \ref{fig:ds_trig}.

The following experimental setup was defined to create the dataset of backdoored networks:

\textbf{Dataset} Casia-Webface \cite{yi_learning_2014} was used. It contains 10,575 identities with a total of 494,414 color images of resolution 250 by 250 in JPEG format.
The dataset is not balanced but was used due to its large number of samples, its sufficient alignment, the grouping per identities and the availability of pretrained models on this dataset in PyTorch. As the dataset does not come with predefined train and test splits, random class splits were used to generate consistent splits across all classes. The exact ratios used were $70\% - 30\%$ for train-test respectively and randomly re-sampled for every experiment.

\textbf{Architecture} The architecture chosen is FaceNet \cite{schroff_facenet_2015}. Implementations can be found online both for TensorFlow\footnote{\url{https://github.com/davidsandberg/facenet}} and for PyTorch\footnote{\url{https://github.com/timesler/facenet-pytorch}}. FaceNet was specifically chosen both for its availability in PyTorch and for having pretrained weights with Casia-Webface. The pretrained weights were used as initialization weights when performing fine-tuning on Casia-WebFace both with and without poisoned samples.

\textbf{Backdoor} The backdoor is a one-to-one with a random impostor-victim pair allowing for the impersonation of one specific person using a digital trigger. The high-level process of both the implementation and activation of the backdoor are shown in figure \ref{fig:backdoor_steps}.

\textbf{Poisoning} The poisoning process involves four main steps shown in figure \ref{fig:backdoor_implementation} (1) the selection of an impostor from the original dataset (involving copying of the samples) (2) the poisoning by application of a trigger on the copied samples (3) the selection of a victim used to relabel the poisoned samples and (4) appending the obtained poisoned samples to the original dataset.

\textbf{Training} The training was performed as for a typical classification, which in the context of biometrics imply a closed-set identification task. The Cross-Entropy loss function was used with increased weights on the impostor and victim classes and the \textit{ADAM} optimizer was selected. We performed the fine-tuning for the clean and the backdoored networks on the \textit{last\_linear} and \textit{last\_bn} layers as the feature extractor is usually best left untouched at the risk of degrading the good generalization performance. Additionally, many biometric tasks are done in open set implying the absence of a classification layer at the end, hence why the focus is on those layers.

\vspace{-0.5cm}
\section{Results}
\label{sec:results}
\vspace{-0.25cm}
The selected layer for our analysis is the \textit{last\_linear} with weights $\pmb{\theta} \in \mathbb{R}^{512 \times 1792}$. In the forward interpretation we consider it as $1792$ vectors of size $512$ while in the backward interpretation we consider the parameters as $512$ vectors of size $1792$. Those selected parameters concatenated from 18 of the clean reference networks first undergo a dimensionality reduction, involving a PCA on the parameters. With the goal of retaining $95\%$ of the cumulative variance, 76 and 297 PCA components are required for the forward and backward interpretation respectively, leading to $18 \times 1792$ vectors of size $76$ in forward and $18 \times 512$ vectors of size $297$ in backward.
The resulting t-SNE visualization (zoomed in for better viewing) is shown in figure \ref{fig:tsne_trig} for the triggers dataset; it can be verified that the parameters for the backdoored networks seem to separate at the edge from the clean parameters.
The optimal number of GMM components can then be determined on the reduced set of parameters. The clusters which can be seen in figure \ref{fig:tsne_trig} suggest that there is a large number of them. An empirical study of the fit of various number of components of the GMM can help in qualitatively evaluating them. For this purpose, the Akaike Information Criterion (AIC) was used, where a score is provided to assess the fit of a GMM on a set of data with the purpose of minimizing the score to identify the optimal number of components. The optimal number of components was identified to be $1792$ and $512$ for forward and backward respectively, out of the following evaluated number of components: [1, 2, 5, 10, 20, 50, 100, 200, 512, 1000, 1792, 3000].

\begin{figure*}[b!]
\centering
\subfloat[Results on the locations dataset.]{\includegraphics[width=0.45\linewidth]{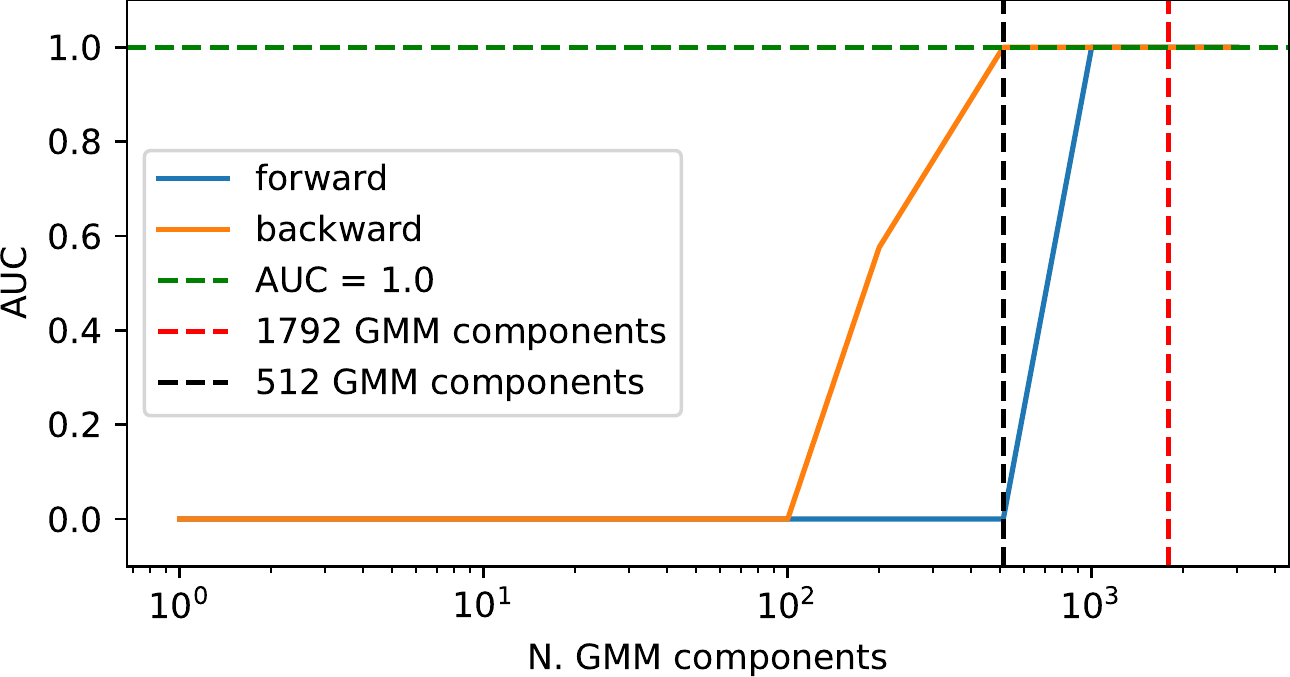}%
\label{fig:auc_loc}}
\hfil
\subfloat[Results on the triggers dataset.]{\includegraphics[width=0.45\linewidth]{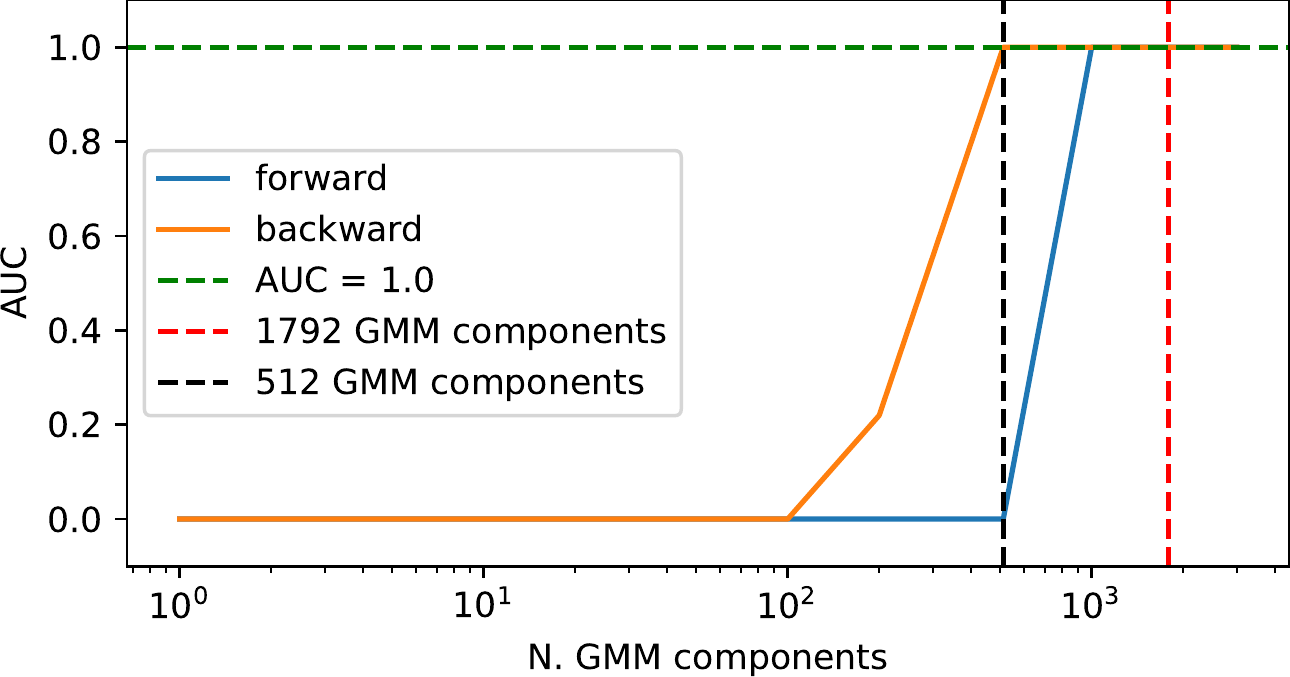}%
\label{fig:auc_trig}}
\caption{AUC for each layer interpretation, as a function of the number of GMM components both for the locations and triggers backdoored datasets in \ref{fig:auc_loc} and \ref{fig:auc_trig} respectively, using the \textit{last\_linear} layer.}
\label{fig:auc}
\end{figure*}

It can be observed that among the evaluated number of GMM components, the optimal number is exactly the number of vectors in the parameters layers (in accordance with the forward and backward interpretation). This suggests that there are indeed in practice a similar distribution of parameters across training experiments, despite using different seeds, leading to the convergence of parameters.

With the GMM fitted and the optimal number of components identified, we test the detector on the remaining unused 12 clean networks and both of the backdoored datasets. We retrieve the scores of each of the networks across both weights representations and evaluate the performance by calculating the the area under the ROC (AUC). The AUC gives one number, summarizing the performance across all thresholds on an ROC. We provide the results of the AUC for all number of GMMs for completeness in figure \ref{fig:auc}. The AUC was used over estimating the threshold due to the need to otherwise split the datasets in three rather than two and it would not be ideally represented with our limited number of networks.
It can be verified that the AUC with our optimal number of gaussians is maximal, with 1.0, across both datasets and both for the forward and backward representations. This implies that there is a perfect threshold on the clean and backdoored networks from the tested networks. Interestingly, the detection in the forward representation seems to work equally well with $1000$ components as with $1792$, which is in line with the AIC making it the second best number of components (not shown). A more in-depth testing may further separate both number of GMM components. On the ability to use this detector in practice, a theoretical false rejection rate (FRR) may be chosen on the clean networks alone, leading to a specific threshold on which performance can be measured.

There are however some limitations to this approach. It requires the training of a certain number of clean networks which can be quite compute intensive. Additionally, one may expect more networks to allow for a better parameters distribution estimation but it may be difficult to estimate a minimum number of networks which may be different on a case-by-case. Lastly, the selection of the layer was chosen in this paper to highlight the feasibility of the method, but in reality the layer is unknown and this method may need to be performed on multiple suspected layers, leading to the use of multiple independent PCA steps and the fitting of a larger number of GMMs.

On the visualization in figure \ref{fig:tsne_trig}, empirically the method works better than what the figure may suggest which could indicate that the visualization may be simplifying to a smaller variability between clean and backdoored networks than what actually is. The t-SNE is a best effort optimization algorithm, not an accurate projection. It attempts at minimizing total loss but it does not allow for absolute distance comparisons. Other projections would also suffer from an inability to accurately visualize relative elements.

\vspace{-0.5cm}
\section{Conclusion}
\label{sec:conclusion}
\vspace{-0.25cm}
The usage of pretrained models hosted online or provided by MLaaS should be possible without needing to blindly trust the hosting entity and service provider: access to convenience features and third parties should not come at the expense of security and we have been able to evaluate a methodology here which allows in the aforementioned conditions to be able to assess whether the networks do in fact come without any hidden backdoor. The pipeline can be automated leading to the selection of the optimal number of PCA components, GMM components, threshold and later be used as binary classification for outlier detection with optimal results with minimal assumptions on the backdoor attack.

This work may also be extended by evaluating detection potential when using another dataset for the same task, possibly leading to the removal of the requirement regarding availability of the clean training set. Additionally, by exploring the optimal method of combining scores when evaluating multiple layers simultaneously. Both of these implementations can improve the generalization capabilities of the methodology and reduce further the number of priors and assumptions if they lead to promising results. The method described in this paper is also expected to perform well in the case of Trajan attacks as the selection of the specific neuron to implement the backdoor will possibly lead to a more significant outlier, which is expected to be easier to identify.

\ifthenelse{\boolean{anonymous}}{}{
\vspace{-0.5cm}
\section{Acknowledgement}
\label{sec:ack}
\vspace{-0.25cm}
This work is supported by the TReSPAsS-ETN project funded from the European Union's Horizon 2020 research and innovation programme under the Marie Skłodowska-Curie grant agreement No.860813.
}

\vspace{-0.5cm}
\bibliographystyle{lnig}
{\footnotesize
    \bibliography{sections/Bibliography}
}

\end{document}